\documentclass{article}
\usepackage{ijcai25}

\usepackage{times}
\usepackage{soul}
\usepackage{url}
\usepackage[hidelinks]{hyperref}
\usepackage[utf8]{inputenc}
\usepackage[small]{caption}
\usepackage{graphicx}
\usepackage{amsmath}
\usepackage{amsthm}
\usepackage{booktabs}
\usepackage{algorithm}
\usepackage{algorithmic}
\usepackage[switch]{lineno}

\newcommand{\citet}[1]{\citeauthor{#1}~[\citeyear{#1}]}

\title{Feature-Guided Neighbor Selection for Non-Expert Evaluation of Model Predictions}

\author{
Courtney Ford$^{1,2}$ \and
Mark T. Keane$^{1,2,3}$ \\
\affiliations
$^1$Research Ireland CRT in Machine Learning, University College Dublin, Ireland\\
$^2$School of Computer Science, University College Dublin, Ireland \\
$^3$Insight Centre for Data Analytics, University College Dublin, Ireland \\
\emails
courtney.ford@ucd.ie
}

\begin{document}

\maketitle

\begin{abstract}
Explainable AI (XAI) methods often struggle to generate clear, interpretable outputs for users without domain expertise. We introduce Feature-Guided Neighbor Selection (FGNS), a post hoc method that enhances interpretability by selecting class-representative examples using both local and global feature importance. In a user study (N = 98) evaluating Kannada script classifications, FGNS significantly improved non-experts' ability to identify model errors while maintaining appropriate agreement with correct predictions. Participants made faster and more accurate decisions compared to those given traditional k-NN explanations. Quantitative analysis shows that FGNS selects neighbors that better reflect class characteristics rather than merely minimizing feature-space distance, leading to more consistent selection and tighter clustering around class prototypes. These results support FGNS as a step toward more human-aligned model assessment, although further work is needed to address the gap between explanation quality and perceived trust\footnote{Accepted at IJCAI 2025 Workshop on User-Aligned Assessment of Adaptive AI Systems}.
\end{abstract}

\section{Introduction}
Explainable Artificial Intelligence (XAI) is crucial in high-stakes domains such as healthcare, finance, and autonomous systems, where AI model decisions can have profound real-world impacts. While methods like Local Interpretable Model-agnostic Explanations (LIME)~\cite{Ribeiro2016} and Shapley Additive Explanations (SHAP)~\cite{Lundberg2017} have made strides in improving transparency, recent work has revealed fundamental limitations in these approaches~\cite{Huang2023,Letoffe2024}. These challenges are particularly evident in misclassification cases, where non-expert users struggle to interpret and validate model decisions~\cite{Ford2023,Knapic2021}.

The gap between domain experts and non-experts is a key challenge in real-world AI deployments~\cite{Jiang2021}. Cognitive science research shows that experts develop superior perceptual processes that aid in analyzing domain-familiar data~\cite{Chase1973}, an advantage that disappears in unfamiliar domains~\cite{Searston2017}. This expertise gap can significantly impact how users interpret AI decisions, especially in image-based systems, with non-experts often misinterpreting model predictions and failing to recognize subtle class boundaries~\cite{Ford2023}.

Existing model-agnostic approaches often assume users can intuitively validate outputs, which can disadvantage those lacking domain expertise~\cite{Szymanski2021}. While recent human-centered XAI approaches have begun addressing this gap~\cite{Ehsan2020}, there remains a critical need for methods that support users with varying levels of domain expertise in understanding and validating AI decisions~\cite{Mueller2019,buccinca2020proxy}. This is especially important in settings where human-in-the-loop assessments are required, but the user may not possess deep knowledge of the underlying task or decision context~\cite{Rudin2019}.

To address these challenges, we introduce Feature-Guided Neighbor Selection (FGNS), a novel XAI method that builds on the twin-system approach~\cite{Kenny2021AIJ}, which combines feature contributions from a convolutional neural network with a k-nearest neighbor (k-NN) classifier for example-based explanations. While this approach has shown promise for familiar domains such as MNIST, non-experts have been shown to struggle when applying these explanations in less familiar contexts~\cite{Ford2023}. FGNS addresses this limitation by integrating local and global feature importance to select class-representative neighbors. It combines LIME's local explanations with Shapley Additive Global ImportancE (SAGE)~\cite{covert2020} global feature importance to identify neighbors that effectively represent class characteristics while remaining relevant to the query instance.

We evaluate FGNS using experiments with Kannada-MNIST, examining its effectiveness in helping non-experts identify model errors and assess prediction confidence compared to traditional example-based explanations. This work contributes to the broader goal of making XAI more accessible to users with varying levels of domain knowledge, particularly in high-stakes domains where accurate validation of AI decisions is critical~\cite{buccinca2020proxy,Severes2023}. Our investigation focuses on both user performance metrics and quantitative measures of explanation quality to understand how feature-guided approaches can help bridge the gap between model complexity and user comprehension.

\section{Related Work}

Explainable AI has made significant strides in increasing the transparency of complex models, yet supporting non-expert users remains a key challenge~\cite{Ridley2025,Severes2023}. This section reviews recent advances in XAI and highlights the limitations of existing methods in supporting users with varying levels of domain expertise.

\subsection{Feature Attribution and Model Transparency}

Despite their popularity, traditional feature attribution approaches face substantial limitations. While LIME~\cite{Ribeiro2016} and SHAP~\cite{Lundberg2017} have become foundational tools in XAI, \citet{Huang2023} highlight key inadequacies in Shapley-based explanations, particularly when applied to complex models. \citet{Letoffe2024} propose refinements to better reflect true feature contributions, while SAGE~\cite{covert2020} extends these methods by providing global importance scores that account for feature interactions across datasets.

The challenge of making such methods accessible to non-technical users has motivated new directions in human-centered XAI~\cite{Wang2019}. Studies show that technical feature importance scores alone are often insufficient for non-experts seeking to understand model behavior~\cite{Knapic2021}, especially in high-stakes settings where reliable interpretation is critical~\cite{buccinca2020proxy}.

\subsection{Prototype and Example-Based Interpretability}

Example-based explanations have emerged as a promising strategy for bridging the gap between model complexity and user understanding. Prototype-based approaches like ProtoPNet~\cite{Chen2019} and Prototype Wrapper Networks~\cite{Kenny2023} enhance interpretability through human-friendly examples, though they frequently assume a degree of domain familiarity that non-experts may not have. This limitation is particularly apparent in unfamiliar classification tasks such as script recognition, where users may not be able to intuitively validate outputs.

Earlier research in case-based reasoning has shown that examples near class boundaries often offer more insight than those selected purely based on proximity to the query~\cite{Doyle2004}. This insight has influenced recent hybrid approaches that combine example selection with feature attribution. For example, \citet{Alfeo2024} bridge local counterfactuals with global importance, while \citet{Lofstrom2024} incorporate uncertainty to generate more calibrated explanations.

Recent work by \citet{DELANEY2023103995} critiques the assumption that plausible counterfactual explanations must be minimally different from the original input. In a user study, they found that humans often produce more substantial edits that align better with class prototypes when asked to explain misclassifications. These findings support the idea that example-based explanations should prioritize semantic clarity and class-representativeness over solely proximity, a principle that FGNS operationalizes through its prototype-guided neighbor selection. 

\subsection{Twin Systems for Enhanced Interpretability}

The twin-system framework~\cite{Kenny2019} offers an alternative route to interpretability by pairing convolutional neural networks with k-nearest neighbor classifiers. One implementation, COLE-HP~\cite{Kenny2021AIJ}, uses the Hadamard product between CNN features and classification weights to select relevant neighbors, making the system’s internal reasoning more transparent.

However, this approach has been shown to present difficulties for non-expert users in less familiar domains~\cite{Ford2023}. These limitations become particularly apparent in misclassification scenarios, where users need to understand decision boundaries to make informed judgments. Such findings underscore the need for methods that better support those without domain expertise, particularly in contexts where class membership is not intuitive~\cite{Jiang2021}.

Ongoing research points toward the need for more inclusive explanation frameworks. For instance, \citet{Ewald2024} propose a rigorous evaluation framework for feature attribution methods in scientific settings, emphasizing the importance of developing techniques that are both robust and interpretable to a wider range of users.

Together, these limitations highlight the need for explanation methods that support post hoc evaluation across diverse domains and user expertise levels. In the next section, we introduce Feature-Guided Neighbor Selection (FGNS), a method designed to improve example-based interpretability by selecting examples that are both class-representative and interpretable to non-expert users.

\section{Methodology}

\begin{figure}[t]
\begin{center}
\includegraphics[width=\columnwidth]{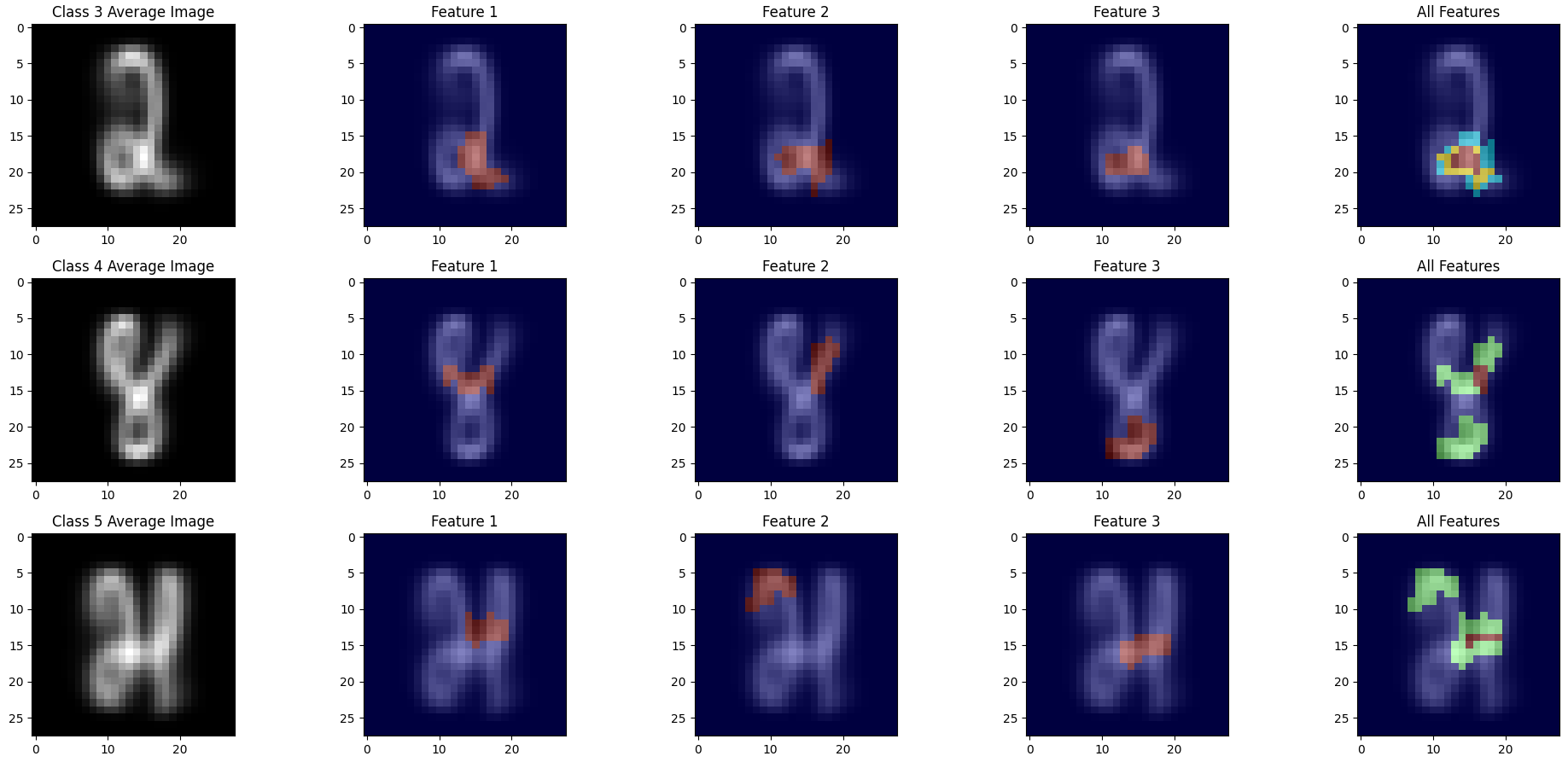}
\caption{Visualization of LIME + SAGE feature integration. For each class: average image (left), individual feature importance masks (middle), and combined feature representation (right), demonstrating how FGNS identifies class-specific features.}
\label{lime-sage-integration}
\end{center}
\end{figure}

Feature-Guided Neighbor Selection (FGNS) is a post hoc explanation method designed to support non-expert users in validating model predictions. It enhances example-based interpretability by selecting neighbors that are not only visually similar, but also aligned with class-representative features. Rather than retraining the model, FGNS re-ranks the neighboring datapoints relative to the query based on their feature-weighted distance to a class prototype, incorporating both local and global feature importance. This results in example selections that better reflect class characteristics and are more representative of prototypical features, particularly in challenging or ambiguous cases.

\subsection{Overview of FGNS Pipeline}
FGNS proceeds in six steps:

\begin{enumerate}
    \item \textbf{Local feature extraction:} For each training image, LIME is used to identify the top-$k$ most important superpixels.
    \item \textbf{Class-level feature aggregation:} For each class, LIME masks are aggregated across 1,000 samples by frequency and spatial overlap. Superpixels that appear frequently or consistently are retained.
    \item \textbf{Global validation:} The aggregated superpixels are evaluated using SAGE, which estimates their contribution to model confidence on class-consistent predictions. Superpixels with low global importance are discarded.
    \item \textbf{Clustering and diversity filtering:} To reduce redundancy, the remaining masks are clustered using K-means ($k=7$) based on their binary shapes. One representative mask is selected per cluster based on SAGE score, and overlapping masks (IoU $\geq$ 0.8) are removed.
    \item \textbf{Prototype construction:} A class prototype is created by computing the pixel-wise median across all training instances from that class.
    \item \textbf{Feature-guided ranking:} For a given test image and predicted class, FGNS computes a loss (Equation~\ref{eq:feature_loss}) between the class prototype and each training instance, weighted by the validated feature masks. The top 3 lowest-loss neighbors are selected as the explanation.
\end{enumerate}

Steps 1-4 are formalized in Algorithm \ref{alg:fgns}, which outlines the LIME + SAGE feature selection process.

\begin{algorithm}[t]
\caption{Feature selection via LIME + SAGE integration}
\label{alg:fgns}
\begin{algorithmic}[1]
\REQUIRE Training data $X$, labels $y$, model $M$
\ENSURE Class feature dictionary $F$
\STATE Initialize empty dictionary $F$
\FOR{each class $c$ in $y$}
    \STATE Sample 1,000 instances from $X_c$
    \FOR{each $x \in X_c$}
        \STATE Apply LIME with 500 perturbations and fixed segmentation to extract top-$k$ superpixels
    \ENDFOR
    \STATE Aggregate superpixels based on frequency and spatial overlap (IoU $\geq$ 0.5)
    \STATE Estimate global importance using SAGE: neutralize each superpixel and measure change in prediction confidence across class samples
    \STATE Retain top masks exceeding global importance threshold
    \STATE Cluster retained masks using K-means ($k=7$); select top-scoring SAGE mask per cluster
    \STATE Remove redundant masks with IoU $\geq$ 0.8
    \STATE Store final set of masks in $F[c]$
\ENDFOR
\STATE \textbf{return } $F$
\end{algorithmic}
\end{algorithm}

\subsection{Prototype Construction}
Each class $c$ is assigned a prototype image $P_c$, computed by taking the pixel-wise median across all training instances:

\begin{equation}
    P_c = \text{median}(\{I_1, I_2, \ldots, I_n\})
\end{equation}

This prototype serves as a reference anchor to evaluate how closely each training instance reflects key class features.

\subsection{Feature-Guided Neighbor Ranking}
Given a test image $I_{\text{query}}$ and its predicted class $c$, FGNS ranks all training instances in class $c$ using a feature-based distance metric:

\begin{equation}
L_{\text{feature}} = \rho \sum_{i=1}^{k} \| M^i \odot (N_{\text{candidate}} - P_c) \|_2^2
\label{eq:feature_loss}
\end{equation}

Here, $M^i$ are the top-$k$ binary feature masks from the LIME+SAGE integration, $\odot$ is the element-wise product, and $\rho$ is a weighting hyperparameter. Lower scores indicate stronger alignment with class-representative features.

FGNS builds on a twin-system setup that uses a CNN for prediction and a k-NN for explanation, following the COLE-HP framework~\cite{Kenny2021AIJ}. After a prediction is made, the k-NN module retrieves candidate neighbors from the predicted class using the Hadamard product of CNN activations and classification weights. FGNS replaces this proximity-based ranking with the feature-weighted loss in Equation~\ref{eq:feature_loss} to prioritize semantically meaningful examples. In our experiments, we set $k=7$ masks (balancing feature diversity with efficiency) and $\rho = 1.0$ (equal mask weighting).

\begin{figure}[t]
\begin{center}
\includegraphics[width=\columnwidth]{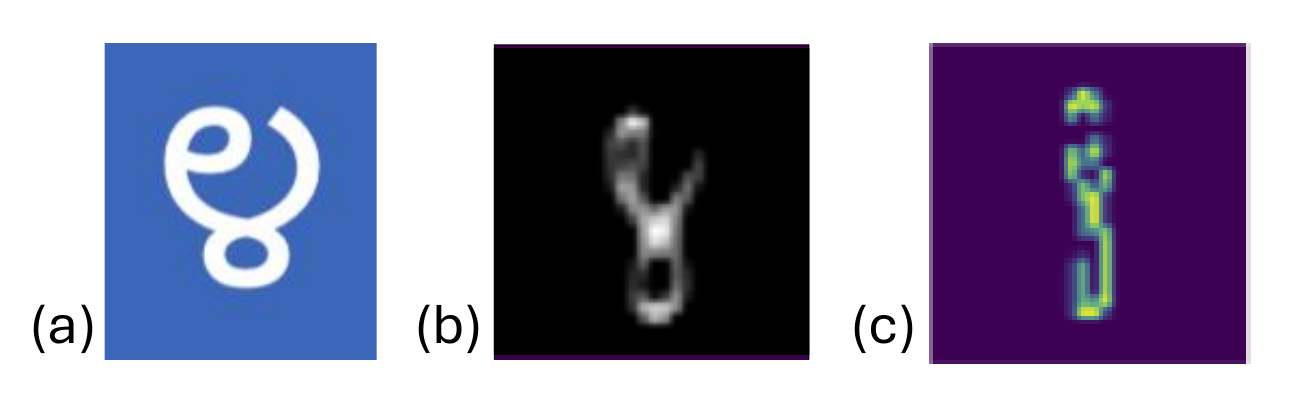}
\caption{Illustration of the Kannada digit "nālku" (4). (a) Ground-truth label; (b) class prototype formed via pixel-wise median; (c) atypical or noisy instance lacking a clear separation between the top curve and end stroke, which are features typically present in the class.}
\label{fig:nalku_example}
\end{center}
\end{figure}

This loss captures how much a candidate neighbor deviates from the prototype in key class-representative regions. To illustrate this, Figure~\ref{fig:nalku_example} shows three Kannada digit images from the "nālku" class: (a) a canonical class label, (b) the prototype computed from the training data, and (c) an atypical instance missing a clear separation between the top curve and end stroke. If (c) were a query image, FGNS would aim to retrieve neighbors that more closely align with (b) by assigning higher loss to candidates that differ from the prototype in class-representative areas.

\begin{figure}[t]
\begin{center}
\includegraphics[width=\columnwidth]{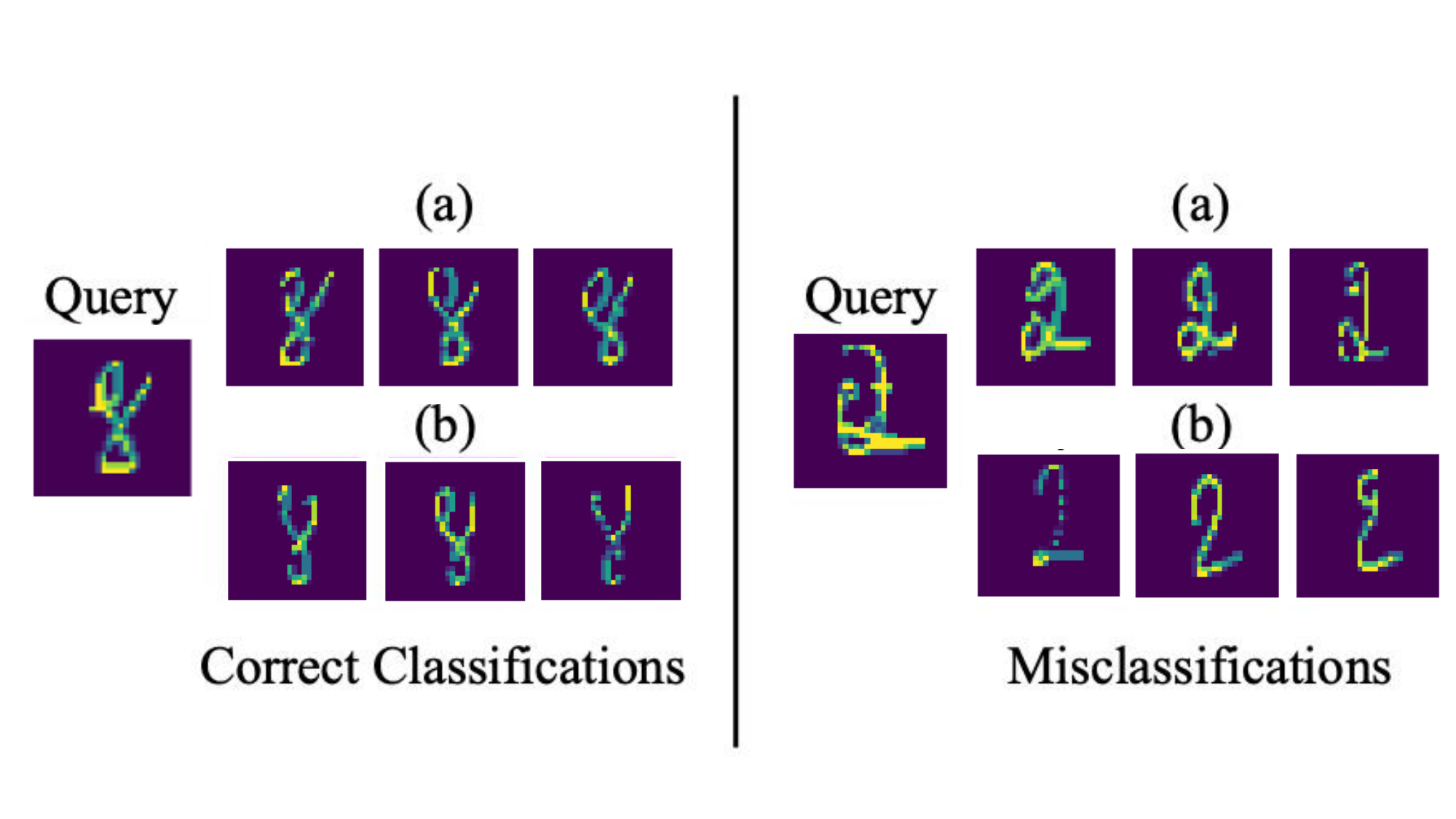}
\caption{Example-based neighboring explanations for correct (left) and misclassified (right) predictions using (a) k-NN and (b) FGNS. Each row shows the query image (left) followed by three nearest neighbors. While both methods return reasonable neighbors for correct predictions, FGNS shows more class-consistent examples, particularly in misclassifications, aiding user interpretation.}
\label{classification_type}
\end{center}
\end{figure}

\subsection{Explanation Generation and Expected Output}
FGNS selects the three lowest-loss training instances as neighbors and presents them alongside the query image. Figure~\ref{classification_type} compares FGNS to traditional k-NN neighbor selection across correct classifications and misclassifications.

For correct classifications (left column), both methods retrieve reasonable neighbors from the same class. For misclassifications (right column), traditional k-NN (row a) selects neighbors based on proximity in feature space, while FGNS (row b) selects neighbors that better represent the predicted class's prototypical characteristics. This makes the mismatch between query and predicted class more apparent and users can see that the query doesn't resemble typical examples of the class the model claims it belongs to.

This approach is designed to help non-experts assess prediction validity through visual comparison, even without domain knowledge of Kannada script.

\section{Evaluation}
We evaluate FGNS through two complementary approaches addressing two research questions: (RQ1) Does FGNS select more class-representative neighbors than proximity-based methods? (RQ2) Do more representative neighbors improve users' error detection abilities and decision efficiency? Our quantitative analysis addresses RQ1, while our controlled user study addresses RQ2. The user study simulates a debugging task where participants must assess prediction correctness using only the query image and explanation examples.

\subsection{Quantitative Analysis}
We conducted three quantitative analyses to evaluate how well FGNS neighbors represented their predicted class, which is a core requirement for reliable post hoc interpretability. Analyses were run on a balanced sample of 50 correctly and 50 incorrectly classified test instances.

\begin{figure*}
\begin{center}
\includegraphics[width=\textwidth]{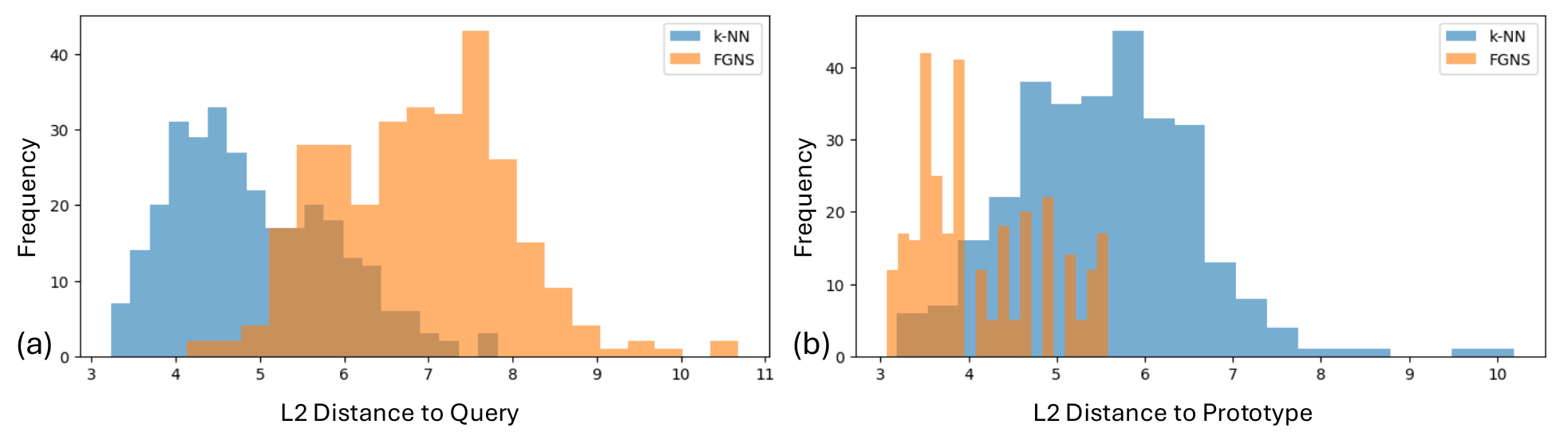}
\caption{Distribution of Euclidean distances for FGNS and k-NN neighbors. (a) Query-to-neighbor distances: FGNS neighbors are farther from the query than those selected by k-NN. (b) Neighbor-to-prototype distances: FGNS neighbors are closer to the predicted class prototype, indicating better semantic alignment.}
\label{l2histogram}
\end{center}
\end{figure*}

\subsubsection{Distance Analysis}
We compared distances to assess whether FGNS prioritizes semantic alignment over spatial proximity. We compared Euclidean (L2) distances between (1) each query and its selected neighbors, and (2) each neighbor and the predicted class prototype. 

FGNS neighbors were significantly further from the query ($M = 6.87$, $IQR = 1.52$) than k-NN neighbors ($M = 4.92$, $IQR = 1.48$), $t(298) = 23.82$, $p < .001$ (Figure~\ref{l2histogram}a). This supports our assumption that FGNS favors semantically aligned rather than spatially proximal instances.

In contrast, when comparing distances to the predicted class prototype, FGNS neighbors were significantly closer ($M = 4.14$, $IQR = 1.21$) than those from k-NN ($M = 5.55$, $IQR = 1.39$), $t(298) = -19.37$, $p < .001$ (Figure~\ref{l2histogram}b). This indicates that FGNS consistently selects examples that better align with the central tendency of the predicted class.

\subsubsection{Cluster Analysis}
We assessed clustering to determine whether FGNS selects more internally consistent examples from each predicted class. We computed the average Euclidean distance between neighbors and their predicted class prototype (cluster centroid), and measured dispersion as the standard deviation of these distances.

FGNS neighbors clustered more tightly around the prototype ($M= 4.14$, $SD = 0.73$) than those from k-NN ($M = 5.55$, $SD = 1.03$). This suggests that FGNS retrieves more internally consistent examples from the predicted class.

\subsubsection{Variance Analysis}
We calculated variance to measure representational consistency, where lower variance indicates tighter class alignment. We calculated the variance of distances between neighbors and the prototype to assess representational consistency.

FGNS neighbors exhibited lower variance ($\sigma^2 = 0.53$) than k-NN ($\sigma^2 = 1.05$), indicating that FGNS more consistently selects examples representative of the predicted class.

\subsection{User Study}
The user study assessed whether FGNS improves participants' ability to understand and evaluate model predictions in a context where they have no prior domain knowledge. Participants were shown model predictions and asked to evaluate their correctness using only the query image and three nearest neighbors, simulating a lightweight debugging task.

\subsubsection{Participants and Design}
We recruited 98 participants through Prolific, targeting native English speakers over the age of 18. Participants were randomly assigned to one of two conditions: the baseline group received traditional k-NN explanations from the COLE-HP algorithm, while the experimental group received FGNS explanations.

\subsubsection{Materials}
We used a CNN-k-NN twin system trained on the Kannada-MNIST dataset, focusing on six number classes (1, 2, 4, 5, 6, 7). These classes were selected for their higher misclassification rates and inter-class confusion, allowing us to examine whether explanations helped participants detect model errors. The test set comprised 36 instances evenly split between correct classifications and misclassifications. Each instance was presented with three neighbor examples selected either by traditional k-NN or FGNS.

\subsubsection{Measures}
We evaluated five dimensions: (1) Classification correctness ratings using a 5-point Likert scale, (2) Judgment accuracy, which was an implicit measure based on whether participants correctly identified the model’s output as right or wrong, (3) Response time, with outliers excluded, (4) Explanation helpfulness using a 5-point Likert scale, and (5) System trust and satisfaction via the DARPA trust and satisfaction questionnaires~\cite{hoffman2018metrics}.

\subsubsection{Procedure}
Participants first received an overview of the task and visual examples of Kannada digits. For each test instance, they viewed the query image alongside its explanation (3 neighbors) and rated agreement with the model's prediction and explanation helpfulness. Finally, they completed the trust and satisfaction survey.

\subsubsection{Classification Correctness}
We measured classification correctness to assess whether users could distinguish correct from incorrect predictions when provided with different explanation types.
A two-way mixed measures ANOVA was conducted to evaluate the effect of Explanation Type (FGNS vs k-NN) and Classification Type (correct vs incorrect) on participants' correctness ratings. The results revealed a significant interaction between Explanation Type and Classification Type, \(F(1, 98) = 9.11, p = 0.003, \eta^2_p = 0.085\). This suggests that the type of explanation provided significantly influenced participants' ability to distinguish correct from incorrect predictions, with a more pronounced effect observed for misclassifications.

Participants who received FGNS explanations rated misclassifications as significantly less correct (\(M = 2.49, SD = 0.52\)) compared to those in the k-NN group (\(M = 2.83, SD = 0.61\)). This indicates that FGNS was more effective in helping users identify errors in the model’s predictions. In contrast, the ratings for correct classifications were nearly identical across the two groups, with FGNS participants rating them at \(M = 4.33, SD = 0.42\) and k-NN participants at \(M = 4.35, SD = 0.45\). This suggests that FGNS did not negatively impact confidence in correct classifications, highlighting its utility in aiding users in distinguishing misclassifications without reducing confidence in the model’s accurate predictions.

\subsubsection{Judgment Accuracy}
We measured judgment accuracy to determine whether users could accurately identify prediction errors, a key requirement for effective model debugging. The two-way mixed measures ANOVA for Judgment Accuracy revealed a significant interaction between Explanation Type and Judgment Type (True vs False), \(F(1, 98) = 7.347, p = 0.008, \eta^2_p = 0.070\), indicating that the type of explanation influenced participants’ ability to make accurate judgments about model predictions. Furthermore, there was a significant main effect of Judgment Type (\(F(1, 98) = 802.214, p < 0.001, \eta^2_p = 0.891\)), showing that participants were generally more accurate in identifying true judgments than false ones.

In this study, True Judgments refer to participants correctly identifying a classification as either correct or incorrect—meaning, they correctly assessed a correct classification as correct and a misclassification as incorrect. On the other hand, False Judgments refer to participants incorrectly identifying a prediction, such as rating a misclassification as correct or a correct classification as an error.

FGNS users demonstrated better performance in True Judgments (\(M = 0.70, SD = 0.11\)) compared to k-NN users (\(M = 0.65, SD = 0.12\)), and they made fewer False Judgments (\(M = 0.17, SD = 0.07\)) compared to k-NN participants (\(M = 0.21, SD = 0.10\)). This shows that FGNS enhanced users’ ability to accurately identify true predictions and reduced the number of false positives. These results suggest that FGNS provided clearer and more intuitive explanations, helping users make more informed and accurate judgments about the model's performance.

\begin{figure}[t]
\begin{center}
\centerline{\includegraphics[width=\columnwidth]{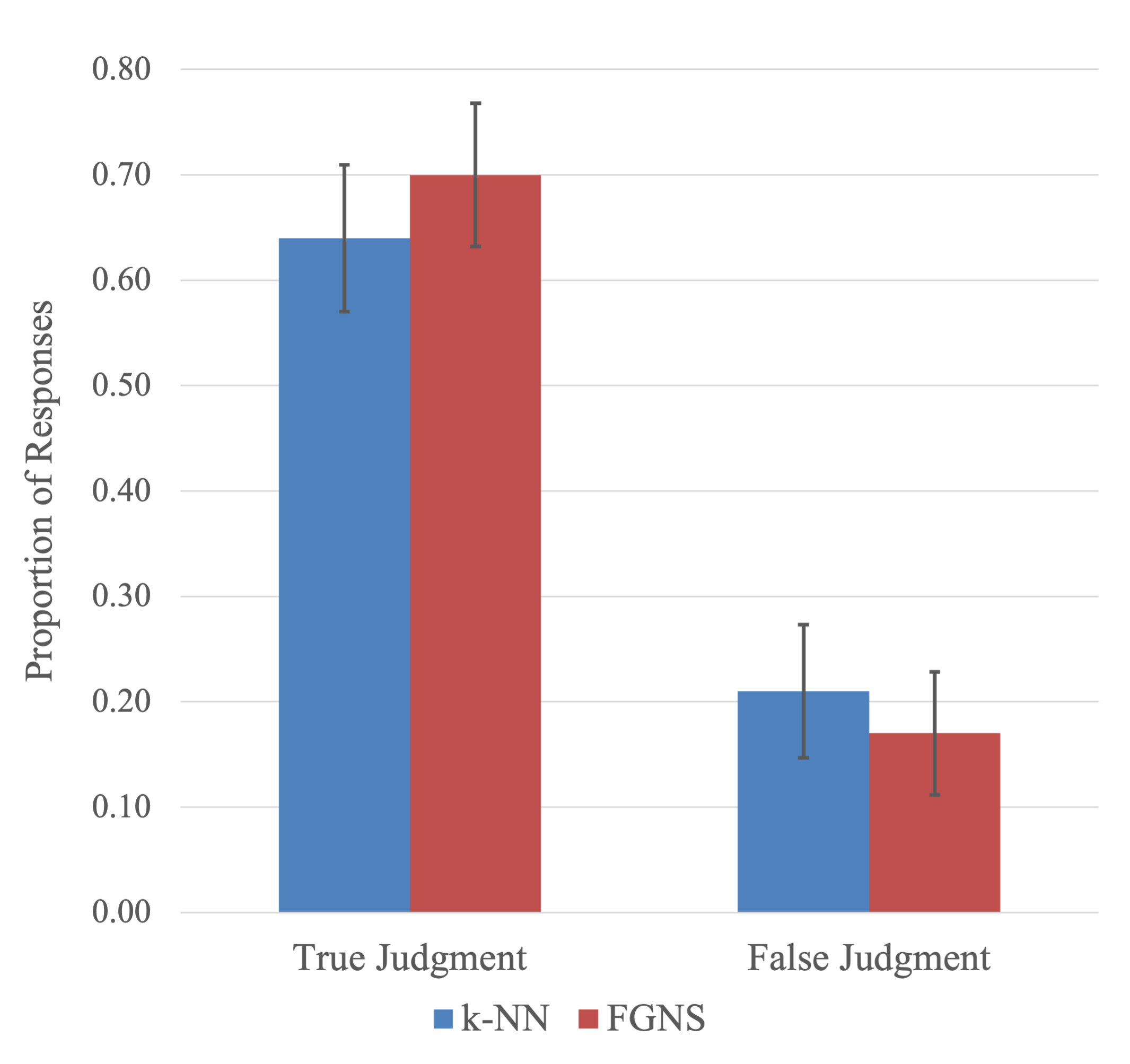}}
\caption{Proportion of True and False Judgments for FGNS and k-NN explanation types. True Judgments refer to correctly identifying a classification as either correct or incorrect. False Judgments refer to misjudging the classification.}
\label{judg_accuracy}
\end{center}
\end{figure}

The proportion of True and False Judgments for both FGNS and k-NN explanation types is shown in Figure~\ref{judg_accuracy}. The graph clearly illustrates the differences between the two explanation types, with FGNS participants showing higher accuracy for True Judgments and lower errors in False Judgments.

\begin{table*}[t]
\centering
\caption{Summary of ANOVA and MANOVA results for user study measures.}
\label{tab:anova_summary}
\begin{tabular}{lllcc}
\toprule
\textbf{Measure} & \textbf{Effect} & \textbf{Statistic} & \textbf{p-value} & \textbf{$\eta^2_p$} \\
\midrule
Correctness Rating 
& Explanation $\times$ Classification & $F(1, 98) = 9.11$ & $0.003$ & $0.085$ \\
\addlinespace[0.5ex]
Judgment Accuracy 
& Explanation $\times$ Judgment Type & $F(1, 98) = 7.35$ & $0.008$ & $0.070$ \\
& Judgment Type & $F(1, 98) = 802.21$ & $< 0.001$ & $0.891$ \\
\addlinespace[0.5ex]
Helpfulness Rating 
& Explanation $\times$ Classification & $F(1, 98) = 0.414$ & $0.521$ & $0.004$ \\
& Classification Type & $F(1, 98) = 300.31$ & $< 0.001$ & $0.754$ \\
\addlinespace[0.5ex]
Response Time 
& Explanation $\times$ Classification & $F(1, 93) = 1.86$ & $0.176$ & $0.020$ \\
& Classification Type & $F(1, 93) = 11.01$ & $0.001$ & $0.106$ \\
& Explanation Type & $F(1, 93) = 4.99$ & $0.028$ & $0.051$ \\
\addlinespace[0.5ex]
Trust (MANOVA) & — & $F(8, 91) = 1.55$ & $0.153$ & $0.12$ \\
Satisfaction (MANOVA) & — & $F(8, 91) = 0.18$ & $0.993$ & $0.015$ \\
\bottomrule
\end{tabular}
\end{table*}

\subsubsection{Response Time}
We measured response time to assess whether clearer explanations impact item-level speed in this debugging task. A two-way mixed ANOVA revealed main effects for both Classification Type (\(F(1, 93) = 11.01, p = 0.001\)) and Explanation Type (\(F(1, 93) = 4.99, p = 0.028\)). FGNS users responded faster overall (\(M = 12.93\)s per item) than k-NN users (\(M = 15.68\)s per item), suggesting that more representative neighbors facilitate quicker decision-making.

\subsubsection{Explanation Helpfulness}
We measured explanation helpfulness to understand users' subjective experience with different neighbor selection methods. A two-way mixed measures ANOVA revealed no significant interaction between Explanation Type and Classification Type (\(p = 0.521\)), suggesting that the type of explanation did not significantly affect the perceived helpfulness for either correct or misclassified predictions. However, a significant main effect of Classification Type was found (\(F(1, 98) = 300.31, p < 0.001, \eta^2_p = 0.754\)), with both FGNS and k-NN explanations rated as more helpful for correct classifications (\(M = 4.33\) to \(M = 4.42\)) than for misclassifications (\(M = 3.11\) to \(M = 3.27\)). The consistently higher helpfulness ratings for correct classifications suggest that users find explanations for correct predictions more intuitive and easier to validate. However, the lower ratings for misclassifications indicate that explaining errors remains a challenging aspect of Explainable AI, even with improvements like FGNS. 

\subsubsection{System Trust and Satisfaction}
We measured trust and satisfaction to assess whether improved explanation quality affects broader user attitudes toward the system. A one-way MANOVA showed no significant differences in trust (\(p = 0.153\)) or satisfaction (\(p = 0.993\)). This aligns with prior work showing that better explanations do not always increase trust~\cite{buccinca2020proxy}. Broader trust may be shaped by prior exposure, expectations, or perceived system transparency, rather than explanation quality alone.

\subsubsection*{Summary of Evaluation Findings}
Table~\ref{tab:anova_summary} summarizes the statistical outcomes from the user study. FGNS significantly improved users’ ability to detect misclassifications, increased overall judgment accuracy, and reduced decision time. It did not, however, significantly affect perceived explanation helpfulness, trust, or satisfaction. These results suggest that while FGNS can enhance specific aspects of decision support, such improvements do not necessarily translate into broader shifts in user perception. This dissociation may be beneficial. FGNS helped users make more accurate judgments without inflating their trust in a system operating in an unfamiliar domain, possibly reducing the risk of overtrust or overreliance on a system that they don't fully understand.

\section{Discussion}

Our findings show that FGNS improves the selection of class-representative neighbors, offering both quantitative and user-level advantages. By prioritizing semantically aligned examples over those selected purely by proximity, FGNS supports more accurate user validation, especially in unfamiliar domains.

Our quantitative results support the central premise of FGNS: that selecting class-representative neighbors leads to more coherent explanations. FGNS neighbors were significantly closer to class prototypes than those from k-NN ($M = 4.14$ vs $5.55$, $p < .001$), indicating stronger alignment with the model’s internal representation of each class. This consistency was further reflected in lower variance and tighter clustering around prototypes, suggesting that FGNS offers a more stable and interpretable explanation space, particularly where class boundaries are visually subtle or ambiguous.

Our user study suggests that FGNS improves non-experts' ability to validate AI decisions, particularly in identifying misclassifications. The significant interaction between explanation type and classification type ($p = 0.003$) revealed that FGNS users rated misclassifications as notably less correct ($M = 2.49$) than k-NN users ($M = 2.83$), while maintaining appropriate confidence in correct predictions (FGNS: $M = 4.33$, k-NN: $M = 4.35$). This selective improvement in error detection without compromising confidence in accurate outputs suggests that FGNS helps users distinguish between correct and erroneous classifications. 

The judgment accuracy results reinforce this interpretation. FGNS users demonstrated higher true judgment rates ($M = 0.70$ vs $M = 0.65$) and lower false judgment rates ($M = 0.17$ vs $M = 0.21$), indicating improved ability to identify correct classifications and reject errors. Notably, users made these judgments more quickly (FGNS: $M = 12.93$s, k-NN: $M = 15.68$s), reinforcing the benefit of principled neighbor selection even in domains unfamiliar to the user.

The integration of LIME and SAGE addresses common pitfalls in feature attribution~\cite{Huang2023,Letoffe2024}, balancing local sensitivity with global stability. For users unfamiliar with a domain, our findings suggest that showing class-consistent rather than merely proximate examples can substantially improve their ability to detect model errors.

However, these performance gains did not translate into higher subjective ratings of explanation helpfulness or overall trust. Misclassification explanations received lower helpfulness ratings (FGNS: $M = 3.11$, k-NN: $M = 3.27$), and there were no significant differences in trust or satisfaction. This aligns with prior findings showing that improvements in task performance do not always affect broader perceptions of AI systems~\cite{buccinca2020proxy}.

\subsection{Limitations}

FGNS was evaluated on a single dataset, Kannada-MNIST, which features clean, centered digit images and limited visual noise. While this choice helped simulate an unfamiliar domain, it may not reflect the complexity of real-world applications. FGNS also relies on interpretable superpixels and well-formed class prototypes, which may be less effective on natural images or in domains without consistent region boundaries. However, existing work suggests that people often prefer explanations that align with prototypical features, even when they require more substantial deviations from the original input~\cite{DELANEY2023103995}. This reinforces the potential of prototype-guided approaches like FGNS to support human-aligned reasoning, though further testing in diverse settings is needed. While our early stopping strategy reduced SAGE runtime, global importance estimation remains computationally intensive. Finally, our evaluation used static explanations; future work could explore FGNS in interactive or time-constrained environments that better reflect real-world deployment.

\section{Conclusion}

We introduced FGNS, a method for selecting class-representative neighbors by integrating local and global feature importance. Across quantitative and user study evaluations, FGNS improved non-experts’ ability to detect model errors without reducing confidence in correct classifications. These results suggest that feature-guided neighbor selection can improve example-based explanations, particularly in domains where users must evaluate predictions without prior expertise.

\subsection{Future Work}

Future work could explore adaptive feature selection, assess explanation diversity or fairness, and extend FGNS to large-scale datasets using approximate search or embedding compression. Most importantly, FGNS should be evaluated in real-world decision support settings such as clinical triage or document review, where transparency and error detection are critical.

\appendix

\section*{Ethical Statement}
This study involved a user experiment with human participants recruited via Prolific. Participants provided informed consent and were compensated fairly. No personally identifiable information was collected. The study received ethics approval from the University College Dublin Research Ethics Committee. No sensitive data were used in the experiments.

\section*{Acknowledgments}
This publication emanated from research conducted with the financial support of Taighde Éireann – Research Ireland under Grant number 18/CRT/6183. For the purpose of Open Access, the author has applied a CC by public copyright licence to any Author Accepted Manuscript version arising from this submission

\bibliographystyle{named}
\bibliography{ijcai25}

\begin{thebibliography}{}

\bibitem[\protect\citeauthoryear{Bu{\c{c}}inca \bgroup \em et al.\egroup }{2020}]{buccinca2020proxy}
Zana Bu{\c{c}}inca, Phoebe Lin, Krzysztof~Z Gajos, and Elena~L Glassman.
\newblock Proxy tasks and subjective measures can be misleading in evaluating explainable ai systems.
\newblock In {\em Proceedings of the 25th International Conference on Intelligent User Interfaces}, pages 454--464, 2020.

\bibitem[\protect\citeauthoryear{Chase and Simon}{1973}]{Chase1973}
William~G. Chase and Herbert~A. Simon.
\newblock Perception in chess.
\newblock {\em Cognitive Psychology}, 4(1):55--81, 1973.

\bibitem[\protect\citeauthoryear{Chen \bgroup \em et al.\egroup }{2019}]{Chen2019}
Chaofan Chen, Oscar Li, Chaofan Tao, Alina~Jade Barnett, Jonathan Su, and Cynthia Rudin.
\newblock This looks like that: Deep learning for interpretable image recognition, 2019.

\bibitem[\protect\citeauthoryear{Covert \bgroup \em et al.\egroup }{2020}]{covert2020}
Ian~C. Covert, Scott Lundberg, and Su-In Lee.
\newblock Understanding global feature contributions with additive importance measures.
\newblock In {\em Proceedings of the 34th International Conference on Neural Information Processing Systems}, NIPS'20, Red Hook, NY, USA, 2020. Curran Associates Inc.

\bibitem[\protect\citeauthoryear{Delaney \bgroup \em et al.\egroup }{2023}]{DELANEY2023103995}
Eoin Delaney, Arjun Pakrashi, Derek Greene, and Mark~T. Keane.
\newblock Counterfactual explanations for misclassified images: How human and machine explanations differ.
\newblock {\em Artificial Intelligence}, 324:103995, 2023.

\bibitem[\protect\citeauthoryear{Doyle \bgroup \em et al.\egroup }{2004}]{Doyle2004}
D{\'o}nal Doyle, P{\'a}draig Cunningham, Derek Bridge, and Yusof Rahman.
\newblock Explanation oriented retrieval.
\newblock In {\em European Conference on Case-Based Reasoning}, pages 157--168. Springer, 2004.

\bibitem[\protect\citeauthoryear{Ehsan and Riedl}{2020}]{Ehsan2020}
Upol Ehsan and Mark~O. Riedl.
\newblock Human-centered explainable ai: Towards a reflective sociotechnical approach.
\newblock In Constantine Stephanidis, Masaaki Kurosu, Helmut Degen, and Lauren Reinerman-Jones, editors, {\em HCI International 2020 - Late Breaking Papers: Multimodality and Intelligence}, pages 449--466, Cham, 2020. Springer International Publishing.

\bibitem[\protect\citeauthoryear{Ewald \bgroup \em et al.\egroup }{2024}]{Ewald2024}
Fiona~Katharina Ewald, Ludwig Bothmann, Marvin~N. Wright, Bernd Bischl, Giuseppe Casalicchio, and Gunnar K{\"o}nig.
\newblock A guide to feature importance methods for scientific inference.
\newblock In Luca Longo, Sebastian Lapuschkin, and Christin Seifert, editors, {\em Explainable Artificial Intelligence}, pages 440--464, Cham, 2024. Springer Nature Switzerland.

\bibitem[\protect\citeauthoryear{Ford and Keane}{2023}]{Ford2023}
Courtney Ford and Mark~T. Keane.
\newblock Explaining classifications to non-experts: An xai user study of post-hoc explanations for a classifier when people lack expertise.
\newblock In Jean-Jacques Rousseau and Bill Kapralos, editors, {\em Pattern Recognition, Computer Vision, and Image Processing. ICPR 2022 International Workshops and Challenges}, pages 246--260, Cham, 2023. Springer Nature Switzerland.

\bibitem[\protect\citeauthoryear{Hoffman \bgroup \em et al.\egroup }{2018}]{hoffman2018metrics}
Robert~R Hoffman, Shane~T Mueller, Gary Klein, and Jordan Litman.
\newblock Metrics for explainable ai: Challenges and prospects.
\newblock {\em arXiv preprint arXiv:1812.04608}, 2018.

\bibitem[\protect\citeauthoryear{Huang and Marques-Silva}{2023}]{Huang2023}
Xuanxiang Huang and Joao Marques-Silva.
\newblock The inadequacy of shapley values for explainability, 2023.

\bibitem[\protect\citeauthoryear{Jiang and Senge}{2021}]{Jiang2021}
Helen Jiang and Erwen Senge.
\newblock On two xai cultures: A case study of non-technical explanations in deployed ai system, 2021.

\bibitem[\protect\citeauthoryear{Kenny and Keane}{2019}]{Kenny2019}
Eoin~M Kenny and Mark~T Keane.
\newblock Twin-systems to explain artificial neural networks using case-based reasoning: comparative tests of feature-weighting methods in ann-cbr twins for xai.
\newblock In {\em Proceedings of the 28th International Joint Conference on Artificial Intelligence}, pages 2708--2715, 2019.

\bibitem[\protect\citeauthoryear{Kenny \bgroup \em et al.\egroup }{2021}]{Kenny2021AIJ}
Eoin~M Kenny, Courtney Ford, Molly Quinn, and Mark~T Keane.
\newblock Explaining black-box classifiers using post-hoc explanations-by-example: The effect of explanations and error-rates in xai user studies.
\newblock {\em Artificial Intelligence}, 294:103459, 2021.

\bibitem[\protect\citeauthoryear{Kenny \bgroup \em et al.\egroup }{2023}]{Kenny2023}
Eoin~M. Kenny, Mycal Tucker, and Julie Shah.
\newblock Towards interpretable deep reinforcement learning with human-friendly prototypes.
\newblock In {\em The Eleventh International Conference on Learning Representations}, 2023.

\bibitem[\protect\citeauthoryear{Knapič \bgroup \em et al.\egroup }{2021}]{Knapic2021}
Samanta Knapič, Avleen Malhi, Rohit Saluja, and Kary Främling.
\newblock Explainable artificial intelligence for human decision support system in the medical domain.
\newblock {\em Machine Learning and Knowledge Extraction}, 3(3):740--770, 2021.

\bibitem[\protect\citeauthoryear{{L. Alfeo} and Cimino}{2024}]{Alfeo2024}
Antonio {L. Alfeo} and Mario Cimino.
\newblock Counterfactual-based feature importance for explainable regression of manufacturing production quality measure.
\newblock In {\em Proceedings of the 13th International Conference on Pattern Recognition Applications and Methods - ICPRAM}, pages 48--56. INSTICC, SciTePress, 2024.

\bibitem[\protect\citeauthoryear{Letoffe \bgroup \em et al.\egroup }{2024}]{Letoffe2024}
Olivier Letoffe, Xuanxiang Huang, Nicholas Asher, and Joao Marques-Silva.
\newblock From shap scores to feature importance scores, 2024.

\bibitem[\protect\citeauthoryear{Lundberg and Lee}{2017}]{Lundberg2017}
Scott~M. Lundberg and Su-In Lee.
\newblock A unified approach to interpreting model predictions.
\newblock In {\em Proceedings of the 31st International Conference on Neural Information Processing Systems}, NIPS'17, page 4768–4777, Red Hook, NY, USA, 2017. Curran Associates Inc.

\bibitem[\protect\citeauthoryear{Löfström \bgroup \em et al.\egroup }{2024}]{Lofstrom2024}
Helena Löfström, Tuwe Löfström, Ulf Johansson, and Cecilia Sönströd.
\newblock Calibrated explanations: With uncertainty information and counterfactuals.
\newblock {\em Expert Systems with Applications}, 246:123154, July 2024.

\bibitem[\protect\citeauthoryear{Mueller \bgroup \em et al.\egroup }{2019}]{Mueller2019}
Shane~T Mueller, Robert~R Hoffman, William Clancey, Abigail Emrey, and Gary Klein.
\newblock Explanation in human-ai systems: A literature meta-review, synopsis of key ideas and publications, and bibliography for explainable ai.
\newblock {\em arXiv preprint arXiv:1902.01876}, 2019.

\bibitem[\protect\citeauthoryear{Ribeiro \bgroup \em et al.\egroup }{2016}]{Ribeiro2016}
Marco~Tulio Ribeiro, Sameer Singh, and Carlos Guestrin.
\newblock "why should i trust you?": Explaining the predictions of any classifier.
\newblock In {\em Proceedings of the 22nd ACM SIGKDD International Conference on Knowledge Discovery and Data Mining}, KDD '16, page 1135–1144, New York, NY, USA, 2016. Association for Computing Machinery.

\bibitem[\protect\citeauthoryear{Ridley}{2025}]{Ridley2025}
Michael Ridley.
\newblock Human-centered explainable artificial intelligence: An annual review of information science and technology (arist) paper.
\newblock {\em Journal of the Association for Information Science and Technology}, 76(1):98--120, 2025.

\bibitem[\protect\citeauthoryear{Rudin}{2019}]{Rudin2019}
Cynthia Rudin.
\newblock Stop explaining black box machine learning models for high stakes decisions and use interpretable models instead.
\newblock {\em Nature Machine Intelligence}, 1(5):206--215, 2019.

\bibitem[\protect\citeauthoryear{Searston and Tangen}{2017}]{Searston2017}
Rachel~A. Searston and Jason~M. Tangen.
\newblock Expertise with unfamiliar objects is flexible to changes in task but not changes in class.
\newblock {\em PLOS ONE}, 12(6):1--14, 06 2017.

\bibitem[\protect\citeauthoryear{Severes \bgroup \em et al.\egroup }{2023}]{Severes2023}
Beatriz Severes, Carolina Carreira, Ana~Beatriz Vieira, Eduardo Gomes, Jo\~{a}o~Tiago Apar\'{\i}cio, and In\^{e}s Pereira.
\newblock The human side of xai: Bridging the gap between ai and non-expert audiences.
\newblock In {\em Proceedings of the 41st ACM International Conference on Design of Communication}, SIGDOC '23, page 126–132, New York, NY, USA, 2023. Association for Computing Machinery.

\bibitem[\protect\citeauthoryear{Szymanski \bgroup \em et al.\egroup }{2021}]{Szymanski2021}
Maxwell Szymanski, Martijn Millecamp, and Katrien Verbert.
\newblock Visual, textual or hybrid: The effect of user expertise on different explanations.
\newblock In {\em 26th International Conference on Intelligent User Interfaces}, IUI '21, page 109–119, New York, NY, USA, 2021. Association for Computing Machinery.

\bibitem[\protect\citeauthoryear{Wang \bgroup \em et al.\egroup }{2019}]{Wang2019}
Danding Wang, Qian Yang, Ashraf Abdul, and Brian~Y. Lim.
\newblock Designing theory-driven user-centric explainable ai.
\newblock In {\em Proceedings of the 2019 CHI Conference on Human Factors in Computing Systems}, CHI '19, page 1–15, New York, NY, USA, 2019. Association for Computing Machinery.

\end{thebibliography}

\end{document}